\pdfoutput=1

\documentclass[11pt]{article}

\usepackage[]{emnlp2021}

\usepackage{times}
\usepackage{latexsym}

\usepackage[T1]{fontenc}

\usepackage[utf8]{inputenc}

\usepackage{microtype}

\usepackage{url}
\usepackage{booktabs}
\usepackage{graphicx}
\usepackage{enumitem}
\usepackage{comment}

\newcommand{\new}[1]{{#1}}
\newcommand{\newer}[1]{{#1}}
\newcommand{\hide}[1]{}

\newcommand{\word}[1] {`\textit{#1}'}
\newcommand{\sense}[1] {`\textsc{{#1}}'}

\usepackage[colorinlistoftodos,prependcaption,textsize=footnotesize]{todonotes}

\title{Grammatical Profiling for Semantic Change Detection}

\author{Mario Giulianelli\thanks{~~Equal contribution, the authors listed alphabetically.} \\
  ILLC,
  University of Amsterdam \\
  \texttt{m.giulianelli@uva.nl} \\\And
  Andrey Kutuzov\footnotemark[1] \\
  University of Oslo \\
  \texttt{andreku@ifi.uio.no}\\\And
  Lidia Pivovarova\footnotemark[1] \\
  University of Helsinki \\
  \texttt{first.last@helsinki.fi}
  }

\begin{document}
\maketitle
\begin{abstract}
Semantics, morphology and syntax are strongly \new{interdependent}.
However, the majority of computational methods for semantic change detection use distributional \new{word} representations \new{which encode} mostly
semantics. We investigate an alternative method,
grammatical profiling, based entirely on changes in the morphosyntactic behaviour of words. We demonstrate that it can be used for semantic change detection \new{and even outperforms some distributional semantic methods.}
\new{We present an} in-depth qualitative and quantitative analysis of \new{the predictions made by our grammatical profiling system},
showing that they are plausible and interpretable.
\end{abstract}

\section{Introduction}

Lexical semantic change detection has recently become a well-represented field in NLP, with several shared tasks conducted for English, German, Latin and Swedish \cite{schlechtweg-etal-2020-semeval}, Italian \cite{basile2020diacr} and Russian \cite{rushifteval2021}. The overwhelming majority of solutions employ either static word embeddings like word2vec \cite{Mikolov_representation:2013} or more recent contextualised language models like ELMo \cite{peters-etal-2018-deep} and BERT \cite{devlin-etal-2019-bert}. These models build upon the distributional semantics hypothesis and can capture lexical meaning, at least to some extent \cite[e.g.,][]{iacobacci2016embeddings,pilehvar2019wic,yenicelik2020bert}.
Thus, they are naturally equipped to model semantic change.

\begin{figure}
    \centering
    \includegraphics[width=1.1\linewidth, trim={0 0 0 1.4cm}, clip]{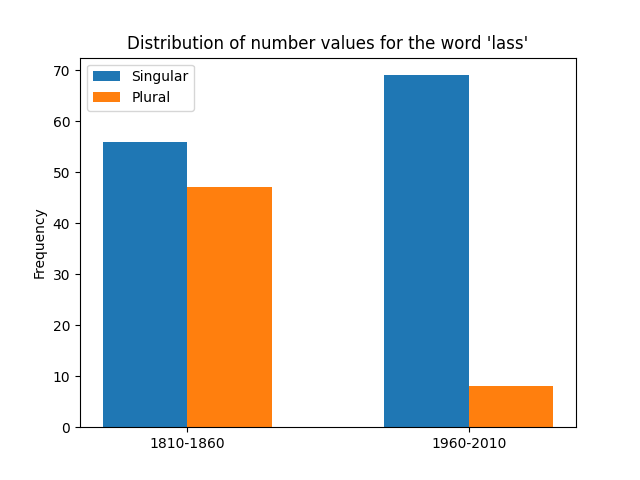}
    \caption{Changes in the number category distribution for the English noun \word{lass} over time, calculated on the English corpora of the SemEval 2020 shared task 1 \cite{schlechtweg-etal-2020-semeval}. \word{Lass} is annotated as semantically changed in the SemEval dataset.}
    \label{fig:lass}
\end{figure}

Yet it has long been known for linguists that semantics, morphology and syntax are strongly interrelated \cite{langacker1987foundations,hock2019language}.
\new{Semantic change is consequently often accompanied by morphosyntactic drifts.}
Consider the English noun \word{lass}: in the 20\textsuperscript{th} century, its \sense{sweetheart} meaning became more dominant over the older sense of \sense{young woman}. This was accompanied by a sharp decrease in plural usages (\word{lasses}), as shown in Figure~\ref{fig:lass}.

Exploiting distributions of \textit{grammatical profiles}---i.e., morphological and syntactic features---to detect lexical semantic change is the focus of this paper.
We investigate to what extent lexical semantic change can be detected using \textit{only} morphosyntax. \newer{Our main hypothesis is that}
significant changes in the distribution of morphosyntactic categories can reveal useful information on the degree of the word's semantic change, even without help from any lexical or explicitly semantic features.

\new{Due to the interdependence of semantics and morphosyntax, it is often difficult to determine which type of change occurred first, and whether it triggered the other.
Establishing the correct causal direction is outside the scope of this study; it is sufficient for us to know that semantic and morphosyntactic changes often co-occur.}

\new{By proposing this functionalist approach to lexical semantic change detection,} we are not aiming at establishing a new state-of-the-art. This is hardly possible without taking semantics into account. But what exactly \textit{is} possible in such a functionalist setup?

We investigate this question experimentally\footnote{\newer{Our code is available at \url{https://github.com/glnmario/semchange-profiling}}
} using standard semantic change datasets for English, German, Swedish, Latin, Italian and Russian.
Our main findings are the following:
\begin{enumerate}[noitemsep,topsep=2pt]
    \item 
    Tracing the changes in the
    distribution of dependency labels, number, case, tense and other morphosyntactic categories outperforms count-based distributional models. In many cases, prediction-based distributional models (static word embeddings) are outperformed as well. This holds across six languages and three different datasets.
    \hide{
    We believe grammatical profiling should be considered as a baseline method for semantic change detection tasks in the future.}
    \item Morphological and syntactic categories are complementary: combining
    them improves semantic change detection performance.
    \item The categories most correlated with semantic change are language-dependent, with number being a good predictor
    cross-linguistically.
    \item The predictions derived from
    grammatical profiling
    are usually interpretable (as in the \word{lass}
    example above), which is not always the case for methods from prior work based on word embeddings, either static or contextualised. This makes our method suitable for linguistic studies that require qualitative explanations.
\end{enumerate}

\section{Related work} \label{sec:rel}
\textit{Behavioural profiles} were introduced in corpus linguistics by~\citet{hanks1996contextual} as the set of syntactic and lexical preferences of a word, revealed by studying a large concordance extracted from a corpus.
The behavioural profile of a word consists of corpus counts of various linguistic properties, including morphological features, preferred types of clauses and phrases, collocates and their semantic types~\cite{griesbehavioral}. Subtle distinctions in word meaning are reflected in behavioural profiles.
Indeed this technique, which combines lexical and grammatical criteria for word sense distinction, was used to study synonymy and polysemy~\cite{divjak2006ways,gries2009behavioral} as well as antonymy~\cite{griesbehavioral}.

One of the theoretical roots for profiling is the theory of \textit{lexical priming}~\cite{hoey2005}. According to this theory, words trigger a set of grammatical and lexical constraints, referred to as \textit{primings} and stored in a mental concordance. The theory states that `Drifts in priming ... provide a mechanism for temporary or permanent language change'~\cite[p. 9]{hoey2005}, and since primings are thought to be organised in the mental concordance in the form of behavioural profiles \cite{griesbehavioral}, it is theoretically plausible that diachronic word meaning change is reflected in a change of behavioural profiles. As far as we are aware, this idea has not been further developed in corpus linguistics.\hide{ Though Gries co-authored several papers related to diachronic corpus comparison~\cite[to name a few]{hilpert2008assessing,hilpert2016quantitative}, the idea of profiles is not explicit in this research.}

In spite of its theoretical validity, behavioural profiling as a practical data analysis technique has serious limitations. Profiles include a large variety of word properties and some of them, especially those related to semantics, cannot be easily extracted from a corpus automatically. Usually, a particular subset of word properties
is selected based on researchers' intuition and background knowledge, and statistical tests are sometimes used for feature selection at later stages of the analysis~\cite{divjak2006ways}.  Moreover, the variety of properties comprised in a behavioural profile makes statistical analysis difficult due to correlations between language phenomena of different levels and sparsity of the data~\cite[section 2.2.2]{kuznetsova2015linguistic}.
For these reasons, some studies~\cite{janda2011grammatical,eckhoff2014grammatical} reduce a word's possibly very broad behavioural profile to a more compact
\textit{grammatical profile}, i.e. a set of preferred morphological forms for the word. These studies too, however, rely on an \textit{a priori} selection of relevant morphological tags.

These technical difficulties may explain
why profiling has not been used in computational approaches to lexical semantic change detection. Most attempts to tackle word meaning change in NLP are based on distributional patterns of \textit{lexical} co-occurrences, starting from early count-based approaches~\cite{juola2003time,hilpert2008assessing},
continuing with dimensionality reduction techniques~\cite{gulordava-baroni-2011-distributional}, 
and later accelerated by embeddings-based models~\cite{kutuzov-etal-2018-diachronic}. More recently, contextualised embeddings were also applied to this task~\cite{giulianelli-etal-2020-analysing,montariol-etal-2021-scalable}.

As far as we are aware, there is one exception to this
trend: \citet{rushifteval_aryzhova} employed grammatical profiles to detect the semantic change of Russian nouns. In their work, a profile of case and number frequency distributions
is collected separately for each time period, and the degree of semantic change is measured as the cosine distance between the two distributions. The results obtained with this method are close to the results yielded by word2vec embeddings, but lower than those of contextualised embeddings.
Inspired by~\citet{rushifteval_aryzhova}, we further investigate the ability of grammatical profiles to capture word meaning change. We propose a number of improvements and evaluate them on datasets in six different languages. Most importantly, we use \textit{all} available morphological tags, without any manual pre-selection, and we conduct an in-depth analysis of our results to understand why grammatical profiling works for this task and what are its limitations.

\section{Data and tasks}
\label{sec:tasks}
\newer{
Following the standard evaluation approach adopted for automatic lexical semantic change detection, we cast the problem as either binary classification (Subtask 1, using the terminology of the SemEval 2020 Unsupervised Lexical Semantic Change Detection shared task \cite{schlechtweg-etal-2020-semeval}) or as a ranking task (Subtask 2). In Subtask 1, given a set of target words, a system must determine whether the words lost or gained any senses between two time periods. In Subtask 2, a system has to rank a set of target words according to the degree of their semantic change.
}

\newer{
Annotating data for word meaning change detection is a non-trivial process because it requires taking into account numerous word occurrences
from every time period of interest. The current practice adopted in the community is to annotate pairs of sentences containing a target word used either in the same or in a different sense; then pairwise scores are aggregated to obtain a final measure of change, either binary or continuous~\cite{schlechtweg-etal-2018-diachronic}. This procedure has been used by organizers of three recent shared tasks: the SemEval 2020 Unsupervised Lexical Semantic Change Detection shared task~\cite{schlechtweg-etal-2020-semeval}, EvaLita \cite{basile2020diacr} and RuShiftEval \cite{rushifteval2021}. We use the data from these three shared tasks, allowing to compare our approach with the state-of-the-art results obtained by
distributional models.
}

\newer{
The SemEval dataset consists of target words in four languages---37 English, 48 German, 40 Latin, and 32 Swedish---that are manually annotated for both subtasks. The EvaLita dataset consists of 18 Italian words annotated for Subtask 1 only. Finally, the RuShiftEval dataset consists of 99 Russian nouns annotated for Subtask 2. All datasets are accompanied by diachronic
corpora. Most of the corpora are split in two time periods, except for the
RuShiftEval corpus, which is separated into
three time bins: \textit{Russian1} and \textit{Russian2} are annotated with semantic shifts between the pre-Soviet and Soviet periods, and between the Soviet and post-Soviet periods respectively; \textit{Russian3} is annotated with semantic shifts between the pre-Soviet and post-Soviet periods~\cite{kutuzov-pivovarova-2021-three}.
}

\newer{In sum, we have at our disposal several dozens words from three Indo-European language groups: Italic, Germanic and Slavic.
Though our results may not generalize to other language families or to other languages within the families analysed, these are the most diverse data that are currently available for this kind of study.
}

\section{Methods}
\label{sec:method}

\subsection{Basic procedure}
\label{sec:baseline}
To obtain grammatical profiles, the target historical corpora are first tagged and parsed with UDPipe \cite{straka-strakova-2017-tokenizing}.\footnote{We use the following models: \textit{english-lines-ud-2.5}, \textit{german-gsd-ud-2.5}, \textit{latin-proiel-ud-2.5}, \textit{swedish-lines-ud-2.5}, \textit{russian-syntagrus-ud-2.5}, \textit{italian-isdt-ud-2.5}.}
Then we count the frequency of morphological and syntactic categories for each target word in both corpora.
More precisely, we count the FEATS values of a corpus's CONLLU file and store the frequencies in two data structures---one for each time period. For example, \texttt{\{'Number=Sing': 338, 'Number=Plur': 114\}} is the morphological dictionary obtained for an English noun in a single time period. We store syntactic features in
an additional dictionary, where keys correspond to the
labels of the dependency arc from the target word to its syntactic head (as found in the DEPREL field of a CONLLU-formatted corpus).

For each target word and for both morphological and syntactic dictionaries, we create a list of features by taking the union of keys in the corresponding dictionaries for the two time bins. The feature list will be \texttt{['Number=Sing', 'Number=Plur']} for the example above.
Then, we create feature vectors $\vec{x}_1$ and $\vec{x}_2$,
where each dimension represents a grammatical category and the value it takes is the frequency of that category in the corresponding time period. If a feature does not occur in a time period, its value is set to $0$. The resulting feature vectors represent grammatical profiles for a word in the corresponding periods. Since the feature list is produced separately for each word, the size of the vectors varies across words.

Finally, we compute the cosine distance $\mathop{cos}(\vec{x}_1, \vec{x}_2)$ between the vectors to quantify the change in the grammatical profiles of the target word.
This is done separately for morphological and syntactic categories, yielding two distance scores $d_{morph}$ and $d_{synt}$. They are used directly to rank words in Subtask 2: the larger is the distance, the stronger is the semantic change.
To solve the binary classification task (Subtask 1), we classify the top $n$ target words in the ranking as `changed' ($1$) and the rest of the list as 'stable' ($0$). The value of $n$ can be either set manually or inferred from the ranking using off-the-shelf algorithms of change point detection \cite{truong2020selective}.

We also combine the scores obtained separately for morphological and syntactic tags by averaging \new{$d_{morph}$ and $d_{synt}$}
for each target word (rounding to the nearest integer in the case of binary classification) and then re-rank the words according to the resulting values. In the end, we have three solutions for each task: `morphology', `syntax' and `averaged'. In the next subsections, we describe a number of improvements that we use to amend this basic procedure.

\subsection{Filtering}
\label{sec:filtering}
To reduce noise that could be introduced due to rare word forms and possible tagging errors,
we exclude rare grammatical categories from the analysis. 
\new{A feature is filtered out from a feature vector $\vec{x}$ if the sum of the feature occurrences in the two time slices amounts to less than five percent of the total word usages. It is possible to optimise this threshold, but we do not tune any numerical parameters to avoid over-fitting to the target datasets.}

\subsection{Category separation}
\label{sec:separation}

In the basic procedure described above,
we extract exactly one morphological feature for each word occurrence; this type of morphological feature is a combination of morphological categories that exhaustively describes a word form.
For example, this is an excerpt from a grammatical profile of the English verb \word{circle} in the 1810-1860 time period:
\vspace{0.3em}\\
\texttt{\small
        \hspace*{0.2cm}Tense=Pres|VerbForm=Part : 50   \\
        \hspace*{0.2cm}Mood=Ind|Tense=Past|VerbForm=Fin : 24 \\
        \hspace*{0.2cm}Tense=Past|VerbForm=Part|Voice=Pass~:~17  \\ 
        \hspace*{0.2cm}VerbForm=Inf : 9 \\
        \hspace*{0.2cm}Mood=Ind|Tense=Pres|VerbForm=Fin : 1 \\
        \hspace*{0.2cm}Tense=Past|VerbForm=Part : 1
} \vspace{0.4em}\\
This representation is very sparse---some features appear only once in the corpus---and it conflates
categories of different nature, such as verb form and tense.
We therefore introduce a category separation step, where feature vectors are created separately for each morphological category. Thus, we transform a distribution of \textit{word forms} into a distribution of \textit{morphological categories} and obtain a denser and more meaningful representation: \vspace{0.3em}\\
\texttt{\small
\hspace*{0.2cm}Tense : \{Past 42, Pres 51\}\\
\hspace*{0.2cm}VerbForm : \{Part 68, Fin 25, Inf 9\}\\
\hspace*{0.2cm}Mood : \{Ind 25\}\\
\hspace*{0.2cm}Voice : \{Pass : 17\}\vspace{0.4em}\\
}
Then cosine distance is computed for each category separately. In the example above, we obtain separate distance values for \texttt{Tense}, \texttt{VerbForm}, \texttt{Mood}, and \texttt{Voice}; the number of distances differs across words and languages.
We take the maximum distance value as the final change score, assuming that a significant change in the distribution of a single category
indicates semantic change, regardless of the other categories.\footnote{We also experimented with averaging  category distances. This improves the results compared to using categories without separation, but it is not as effective as taking the maximum.}

When separation is combined with filtering, filtering is performed \textit{after} feature separation to preserve maximum information.
Continuing with the previous example: in the basic procedure, the word form \texttt{Tense=Past|VerbForm=Part} is filtered out, as \new{it appears once in the first corpus and} it is rare in the second corpus as well. In the category separation strategy this form is taken into account, separately contributing to the \texttt{Tense} and \texttt{VerbForm} distances.

\subsection{Combination of morphology and syntax}
\label{sec:method-combination}
Category separation opens new possibilities for taking syntactic categories into account. We can average morphological and syntactic distances, as in our basic procedure, or append the syntactic distance value to the array of morphological distances, and then choose the maximum. In the first strategy, morphological and syntactic rankings are weighted equally regardless of the number of morphological categories for a given word. In the second strategy, syntactic labels are weighted \new{down} depending on the richness of the morphological profile.

\section{Results}
\label{sec:results}

We evaluate our method on both subtasks of the SemEval 2020 Unsupervised Lexical Semantic Change Detection shared task \cite{schlechtweg-etal-2020-semeval}. As described in Section~\ref{sec:tasks}, Subtask 1 is a binary classification task, evaluated with accuracy.
Subtask 2 is a ranking task, evaluated with Spearman’s rank correlation.

\begin{table*}
\centering
\resizebox{\textwidth}{!}{  
    \begin{tabular}{l|ccccc|cccc}
    \toprule
\textbf{Categories} &\multicolumn{5}{c}{\textbf{SemEval 2020 languages}}     & \multicolumn{4}{c}{\textbf{Russian}}   \\
\midrule
& English  & German & Latin &	Swedish &	Mean     &  Russian1  &  Russian2 &  Russian3 & Mean \\

\midrule
&\multicolumn{9}{c}{\textbf{Basic procedure}} \\
\midrule
Morphology &   0.234 &	0.043 &	0.241 &	0.207 &	0.181  &  \textbf{0.137}   & 0.210    & \textbf{0.327}    & \textbf{0.225}\\
Syntax     &   0.319 &	0.163 &	0.328 &	-0.017 & 0.198 &  0.060   & 0.101  & 0.269    & 0.143\\
Average    &   0.293 &	0.147 &	0.304 &	0.088 &	0.208  &   0.101   & 0.191  & 0.294    & 0.195\\
\midrule
& \multicolumn{9}{c}{\textbf{5\% filtering}} \\
\midrule
Morphology   &   0.211 &  0.080 & 0.285 &   0.191 & 0.192     & 0.127   & 0.185    & 0.264    & 0.192 \\
Syntax       &   \textbf{0.331} &  0.146 & 0.265 &   0.184 & 0.231  & 0.056   & 0.111    & 0.279    & 0.149 \\
Average      &   0.315 &  0.171 & 0.345 &   0.263 & 0.273    & 0.094   & 0.183    & 0.278    & 0.185 \\
\midrule
& \multicolumn{9}{c}{\textbf{Category separation and 5\% filtering}} \\
\midrule
 Morphology   &  0.218 &  0.074 & 0.519 & 0.303  & 0.278                             &  0.028    & \textbf{0.241}    & 0.293    & 0.187 \\
 Average      &  0.321 &  0.227 & 0.523 &  \textbf{0.381} & 0.363 		     &  0.002    & 0.179    & 0.278    & 0.153 \\
Combination  &  0.320 &  \textbf{0.298} & \textbf{0.525} &  0.334 & \textbf{0.369}  &  0.000    & 0.149    & 0.242    & 0.130 \\
\midrule
\midrule
    \multicolumn{6}{c|}{\textbf{Prior SemEval results}} &     \multicolumn{4}{c}{\textbf{Prior RuShiftEval results*}} \\
    \midrule
    Count baseline & 0.022 &	0.216 &	0.359 &	-0.022 &	0.144  & 0.314  & 0.302    & 0.381    &     0.332  \\
    Best shared task system & 0.422 &	0.725	& 0.412 &	0.547 &	0.527 &  0.798   & 0.803    & 0.822    &     0.807  \\
\cite{rushifteval_aryzhova} &-&-&-&-& -& 0.157   & 0.199    & 0.343    & 0.233  \\
    \bottomrule
    \end{tabular}
    }
    \caption{Performance in graded change detection (SemEval'20 Subtask 2 and RuShiftEval),  Spearman rank correlation coefficients. Note that RuShiftEval features three test sets for three different time period pairs.  \\
    \footnotesize{*The RuShiftEval baseline relies on CBOW word embeddings and their local neighborhood similarity. \cite{rushifteval_aryzhova} used an ensemble method with much higher performance, we report the results obtained solely with profiling. While SemEval results are fully unsupervised, the best RuShiftEval results are supervised and not directly comparable to our setting.}}
    \label{tab:graded}
\end{table*}

\paragraph{Basic procedure} Using only morphological features, we obtain an average correlation of $0.181$ across the four SemEval languages, as can be seen in Table~\ref{tab:graded}. Syntactic features yield a $+0.017$ increase, and after averaging $d_{morph}$ and $d_{synt}$ (see Section \ref{sec:baseline}) we reach a correlation score of $0.208$. This is already substantially higher than the SemEval baseline which employed
count-based distributional models (see Table \ref{tab:graded}).

\paragraph{Frequency threshold} Filtering out rare features as described in Section \ref{sec:filtering} has a \new{small but} positive impact on all three setups: $+0.011$ for morphological features, $+0.033$ for syntactic features, and $+0.065$ for the combination of the two.

\paragraph{Category separation} Measuring distance between morphological categories separately (see Section \ref{sec:separation}) produces an additional significant boost: we obtain a correlation score of $0.278$ using
these refined \new{morphological} representations. In combination with syntactic features (Section~\ref{sec:method-combination}), this approach yields an average correlation of $0.369$ with human judgements. This is our best result on Subtask 2,  more than twice higher than a correlation obtained by the SemEval count-based baseline (see Table \ref{tab:graded}); for Latin, a language with rich morphology, grammatical profiles actually outperform even the \textit{best} SemEval 2020 submission. These scores are particularly impressive given that, unlike \new{those based on} distributional vectors, our method has no access to lexical semantic information.

As can be seen in Table~\ref{tab:graded}, \new{our category separation approach}
does not extend well to the Russian test sets, obtaining an average correlation score of $0.130$.\footnote{At the same time, in the basic procedure,  morphological features yield a much higher correlation score of $0.225$.}
A possible explanation for the lower correlation may be related to smaller distances between Russian time bins as compared to the SemEval setup: \textit{Russian1} and \textit{Russian2} are annotated with semantic shifts between pre-Soviet and Soviet and between Soviet and post-Soviet periods respectively, while \textit{Russian3} measures the change between pre-Soviet and post-Soviet periods, with a significant time gap in between. Indeed we obtain much higher scores on \textit{Russian3}. In addition, the annotation procedures for the RuShiftEval dataset differ in some details from those for SemEval'20.

Another observation is that \new{morphological} category separation does not improve results for English. The best method for English relies only on syntactic features.
The most plausible explanation is that English \new{morphology is rather poor and it tends}
to mark grammatical \new{categories}
with separate words. Our method can be potentially improved by taking into account multi-word forms, e.g.\ to determine English verb mood.

\begin{table*}
\centering
\footnotesize
    \begin{tabular}{l|ccccc|c}
    \toprule
    \textbf{Categories} & \textbf{English}  & \textbf{German} & \textbf{Latin} &	\textbf{Swedish} &	\textbf{Mean} & \textbf{Italian}\\
     \midrule
&\multicolumn{6}{c}{\textbf{Basic procedure}} \\
\midrule
    Morphology  &   0.595 &	0.521 &	0.525 &	0.581 &	0.555 & 0.722 \\
    Syntax    & 0.541 &	\textbf{0.646} &	0.575 &	0.645 &	0.602 & 0.611 \\
    Average    & 0.568 &	0.583 &	0.475 &	\textbf{0.710} &	0.584 & 0.722 \\
    \midrule
    &\multicolumn{6}{c}{\textbf{Automatic change point detection}} \\
    \midrule
    Morphology   &   \textbf{0.622} &  0.479 & \textbf{0.625} &   0.548 &0.569 & 0.722   \\
    Syntax       &   0.514 &  0.625 & 0.500 &   0.677 &0.579 & 0.611  \\
    Average      &   0.595 &  0.542 & 0.525 &   0.677 &0.585 & \textbf{0.778}  \\
 \midrule
    &\multicolumn{6}{c}{\textbf{Category separation, change point detection and 5\% filtering}} \\
    \midrule    
Morphology  &    0.622 &  0.583 & \textbf{0.625} &   0.581 &\textbf{0.603} & 0.500\\
Average      &   0.595 &  0.625 & 0.450 &   \textbf{0.710} &0.595 & 0.667\\
Combination  &   0.541 &  0.583 & 0.575 &   0.645 &0.586 & 0.500\\
    \midrule
    \midrule
    \multicolumn{6}{c|}{\textbf{Prior SemEval results}} & \multicolumn{1}{c}{\textbf{Prior EvaLita results*}} \\
    \midrule
    Baseline &0.595 &	0.688	& 0.525 &	0.645 &	0.613 & 0.611 \\
    Best shared task system & 0.622	& 0.750 &	0.700 &	0.677 &	0.687 & 0.944 \\
    \bottomrule
    \end{tabular}
    \caption{Performance in binary change detection (SemEval'20 Subtask 1 and EvaLita), accuracy. Note that in this paper we mostly focus on ranking (Subtask 2). All the binary change detection methods here are entirely based on the scores produces by the ranking methods.   \\
    \footnotesize{*The Italian baseline relies on collocations \cite{basile-etal-2019-diachronic}: for each target word, two vector representations are built, with the Bag-of-Collocations related to the two different time periods. Then, the cosine similarity between them is computed.}}
    \label{tab:binary}
\end{table*}

\paragraph{Subtask 1} 
Following our basic procedure (Section \ref{sec:baseline}), we assign a classification score of 1 to the top 43\% of the target words\footnote{Average ratio of changed words across SemEval datasets.} for each language, ranked according to their grammatical profile changes. This yields an accuracy close to that of the SemEval count-based baseline (see Table \ref{tab:binary}).\footnote{Note that the SemEval'20 count baseline also uses a manually defined threshold value in Subtask 1.}
Filtering rare features hardly yields any improvement here, but once combined with morphological category separation and automatic change point detection it produces an accuracy of $0.603$.
We also observe that using change point detection with dynamic programming \cite{truong2020selective} does not cause any significant accuracy decrease in comparison to using the hard-coded 43\% ratio, showing that our method does not require knowledge of the test data distribution.
On the Italian test set, we correctly classify 3 more words (out of 18) than the collocation-based baseline \cite{basile2019kronos}, obtaining an accuracy of $0.778$.

\section{Qualitative analysis} 
\label{sec:analysis}
In Section \ref{sec:results}, we showed that grammatical profiling alone can detect a word meaning change better than count-based distributional semantic models which exploit lexical co-occurrence statistics. This is a remarkable finding: it confirms that meaning change leaves traces in grammatical profiles and it demonstrates that these traces can be used as effective predictors of a word's meaning stability.
In this Section, to better understand when change in grammatical profiles is a good indicator of lexical semantic change, we analyse the characteristics of the target words to which our method assigns the most and least accurate rankings.

\subsection{When is grammatical profiling enough?}
We begin by analysing the most accurately ranked words (see Appendix~\ref{sec:app-predictions}).
The Italian word \word{lucciola}, for example, is ranked 1\textsuperscript{st} out of 18 by our method due to the singular usages of the word disappearing after 1990. The singular usage is indeed much more likely for the dying sense of the word (an euphemism for \sense{prostitute}), whereas the plural form \word{lucciole} is more likely used for the stable sense of the word (\sense{fireflies}) or in the idiomatic expression \textit{prendere lucciole per lanterne} (\textit{getting the wrong end of the stick}), which makes up for most of the occurrences between 1990 and 2014. Another example of correctly identified semantically shifted words is the Latin \word{imperator} (ranked 1\textsuperscript{st} out of 40). In the second time period---ranging from 0 to 2000 A.D.---nominative usages become predominant. \new{A possible explanation for this change is that the more frequent agentive usages of the word correspond}
to the new role of the \sense{emperor} in the imperial Rome (27 B.C. to A.D. 476)
rather than that of a \new{generic} \sense{commander}---the older sense of the word.\footnote{\new{We are aware that the current separation of the Latin corpus into two time periods can be controversial. Still, we follow the splits defined by the SemEval 2020 organisers \cite{schlechtweg-etal-2020-semeval} for consistency and comparability with prior work.}}

For English, the noun \word{stab} is ranked 4\textsuperscript{th} out of 37, mostly because of syntactic changes: 27\% of its occurrences in the 20\textsuperscript{th} century are used as oblique arguments, compared to only 13\% in the 19\textsuperscript{th} century. This is arguably associated with the emergent sense of \sense{sudden sharp feeling} (\word{...left me with a sharp stab of sadness}).
The German word \word{artikulieren} correctly receives a high rank (9\textsuperscript{th} out of 48): it occurs only 3 times in the 19th century and 210 times between 1946 and 1990, shifting towards a much richer grammatical profile. Sharp changes in frequency are reflected in the diversity of grammatical profiles and can also help detect lexical semantic change.

\new{Our qualitative analysis reveals that the successful examples are often cases of broadening and narrowing of word meaning. These kinds of semantic change seem to be easily picked with profiling. However, some examples of broadening and narrowing fail to be detected, as will be shown in Section~\ref{sec:false-pos-neg}, especially if they involve metaphorical extensions of word meaning. \hide{We analyse such cases as \textit{false negatives} in the next Section. }A consistent characterisation of the kinds of semantic change detected and overlooked by our method would require diachronic corpora where both the degree and the type of semantic changes are annotated. }

\subsection{When it is not enough?}
\label{sec:false-pos-neg}
Although it largely outperforms simple distributional semantic models, our grammatical profiling approach is still not on par with state-of-the-art semantics-based algorithms. To find out when changes in morphosyntactic profiles are not sufficient to detect a word's meaning change, we analyse \textit{false positives} and \textit{false negatives}: i.e., target words that are assigned an erroneously high or low semantic change score, respectively.

\textbf{False positives} are words whose change in grammatical profile does not correspond to semantic change. An example of a false positive is the Italian word \word{cappuccio} (\sense{hood}). The increase from 9\% to 41\% of plural usages causes our method to assign this word a relatively high change score---6\textsuperscript{th} out of 18 (6 words are annotated as changing in the Italian dataset). Inspecting the Italian corpora, we notice that between 1945 and 1970 the word is mainly used to describe the pointed hood of the robes typically worn by Ku Klux Klan members; after 1990, the word's context of usage becomes much less narrow. The meaning of the word, however, does not change. \new{This type of errors is, at least to a certain extent, an artifact of the source data: grammatical profiles are less accurate when the set of domains covered by a corpus is limited.}

\new{Another type of false positives is also partially related to corpus imbalance.}
We have seen in the previous section that sharp frequency increases correspond to significant changes in grammatical profiles, and that this information can be exploited by our method to detect changing words.
However, frequency change can be an unfaithful indicator of meaning change. This is the case, for example, for the German words \word{Lyzeum} (\sense{lyceum}; ranked 1\textsuperscript{st} out of 48), and \word{Truppenteil} (a \sense{unit of troops}; ranked 11\textsuperscript{th}),
and for the Latin word \word{jus} (a \sense{right}, the \sense{law}; ranked 4\textsuperscript{th} out of 40).

\textbf{False negatives}, on the other hand, are words whose semantic change is not reflected in changes in grammatical profile.
The German word \word{ausspannen} (\sense{to remove}, \sense{to unclamp}) is used across the 19\textsuperscript{th} and 20\textsuperscript{th} century
only in its infinitive form, so our method assigns it a relatively low change score (23\textsuperscript{rd} out of 48). Most of the occurrences in the 19\textsuperscript{th} century, however, are literal usages of the word (e.g., \textit{die Pferde ausspannen}, \textit{to unhitch the horses}), whereas in the (second part of the) 20\textsuperscript{th} century the novel metaphorical usage of the word (e.g., \textit{für fünf Minuten ausspannen}, \textit{to relax for five minutes}) is the most frequent one. Another example of a German word whose novel metaphorical sense remains undetected (ranked 31\textsuperscript{st}) is \word{Ohrwurm} (\sense{earworm}): the grammatical profile of this word remains stable (except for the accusative case becoming slighlty more frequent), but the word acquires the meaning of \textit{catchy song}, or \textit{haunting melody}.
Similarly, the singular usages of the Latin word \word{pontifex} increase from 63\% to 83\%, signalling the semantic narrowing of the word occurred in medieval Latin (from a \sense{bishop} to the \sense{Pope}), but the case distribution remains similar; this results in a rather low change score (ranked 22\textsuperscript{nd} out of 40).
The last two examples show that taking the maximum distance across categories (see \ref{sec:separation}) is a correct strategy, yet sometimes the changes in that grammatical category are still insufficient for our method to detect change.

\section{Category importance} 
\label{sec:feature}

In this Section, we conduct an additional experiment to find out which grammatical categories are most related to semantic change. To this end, we train logistic regression classifiers for binary classification using English, German, Latin, Swedish and Italian data. The classifier features are cosine distances between frequency vectors of each particular category from different time bins. Before fitting the classifier, each feature is independently zero-centered and scaled to the unit variance.
Then, regression coefficients are estimated for each feature: we consider positive weights as an indication of usefulness of a feature for classification. The outcome of this analysis is shown in Table~\ref{tab:features_subtask1}. We list English nouns and verbs separately since the SemEval'20 dataset explicitly annotates part-of-speech tags for the English target words. This is not the case for the other languages in this dataset.

In line with the results presented in Section~\ref{sec:results}, Swedish and Italian classifiers yield the highest accuracy and F-score. Latin, a highly inflectional language, has by far the largest set of categories contributing positively to semantic change detection (interestingly, excluding syntax). English, a highly analytical language, is on the other end of the spectrum.

\begin{table}
    \centering
    \resizebox{\columnwidth}{!}{  
    \begin{tabular}{l|p{0.4\columnwidth}|ll}
    \toprule
    \textbf{Language} & \textbf{Top categories} & \textbf{Accur.} & \textbf{F1} \\
    \midrule
    \textbf{English nouns} & number & 0.576 & 0.523  \\
    \textbf{English verbs} & verb form, syntax  & 0.750 & 0.733  \\
    \midrule
    \textbf{German} & number, syntax, gender  & 0.542 & 0.541  \\
    \midrule
    \textbf{Swedish} & syntax, mood, voice, definiteness, number & 0.839 & 0.797  \\
    \midrule
    \textbf{Latin} & voice, number, degree, case, gender, mood, aspect, person, tense & 0.650 & 0.649  \\
    \midrule
    \textbf{Italian} & number, tense, syntax & 0.778 & 0.723 \\
    \bottomrule
    \end{tabular}
    }
    \caption{Categories with positive weights in binary classifiers of semantic change (logistic regression). `Syntax' stands for dependency relation to the syntactic head of the word. Evaluation scores are calculated on the train data, F1 is macro-averaged.}
    \label{tab:features_subtask1}
\end{table}

Additionally, we estimate the relative importance of morphosyntactic categories by calculating the Spearman's rank-correlation of their respective cosine distance values (across all target words) with the gold semantic change rankings. In other words, we single out each category, e.g.\ verbal mood, and test whether diachronic change in its frequency distribution is correlated with manually annotated semantic change scores.

In Table~\ref{tab:features_subtask2}, we show the categories with statistically significant ($p < 0.05$) correlations for each language and dataset.  In English, as expected given its analytical nature, only changes in syntactic roles yield such a correlation; other categories are either non-existent in this language, or are not linked to semantic change strongly enough. \new{For an inflection language such as} 
Latin, number and adjectival degree are highly predictive (the latter is arguably because Latin has the highest ratio of adjectives among all SemEval 2020 Task 1 datasets: about 20\%).
\begin{table}
    \centering
    \resizebox{\columnwidth}{!}{
    \begin{tabular}{l|cccccc}
        &\textbf{Number} &\textbf{Mood} &\textbf{Degree} &\textbf{Gender} &\textbf{Case} &\textbf{Syntax}  \\
         \midrule
         \textbf{English} & - &- &- &- &-& 0.331\\
         \textbf{German} & - & -& -& -& -&- \\
         \textbf{Latin} & 0.304 &- &0.301 &- &- &- \\
         \textbf{Swedish} & 0.402 & 0.397 &- &- &- &- \\
         \midrule
         \textbf{Russian 1} &- &- &- &0.218 &0.196 &- \\
         \textbf{Russian 2} &- &- &- &0.231 &0.324 &- \\
         \textbf{Russian 3} & 0.246 &- &- &0.218 & 0.327 & 0.279 \\
         \bottomrule
    \end{tabular}
    }
  \caption{Spearman rank correlations between diachronic grammatical profile distances for different categories and manually annotated semantic change estimations. `-' stands for no significant correlation.}
    \label{tab:features_subtask2}
\end{table}
Not surprisingly for a synthetic language, the morphological categories of number and case show strong correlations for Russian. In the case of the larger time gap between pre-Soviet and post-Soviet periods (Russian 3), syntactic relationships also become a good predictor.

What \textit{is} surprising, however, is that changes in gender are also correlated with semantic change in the Russian case. This result is hard to interpret,
since grammatical gender is a lexical feature of Russian nouns and does not change from occurrence to occurrence;  even diachronically, such cases are quite rare. The reason for this is slightly erroneous morphological tagging: our tagger mixes up homographic inflected forms, which abound in Russian, and assigns feminine gender to masculine nouns, and vice versa. The reliance on the tagger performance can be seen as a limitation of our grammatical profiling approach. However, the existence of the correlation hints that these errors are not entirely random, and their frequency is influenced by word usage: gender is ambiguous only in certain case and number combinations, and \new{the frequency of these combinations}
seems to change diachronically. For example, for the form \word{cheki} \new{(}`cheques/grenade pin'\new{)}, the masculine lemma licenses the accusative plural reading, while the feminine lemma licenses the genitive singular reading. Thus, even the tagger errors are in fact informative.

Interestingly, for German, no single category changes are significantly correlated with semantic change. This is in line with our weak---although still higher than the count-based baseline---results for German described above, but is somewhat surprising, given the fusional nature of the language, with its rich spectrum of inflected word forms.\footnote{\newer{We computed correlations for German nouns and verbs separately, but did not find any significant correlation either.}} Some peculiarities of the employed tagger model might be responsible for this finding, which should be further tested and explained in future work.

\section{Conclusion} \label{sec:conclusion}
Semantic change is inextricably tied to changes in the distribution of morphosyntactic properties of words, i.e. their grammatical profiles. In this paper, we showed that tracking these changes is enough to build a semantic change detection system which, without access to any lexical semantic information, consistently outperforms count-based distributional semantic approaches to the task.
Grammatical profiling yields surprisingly good evaluation scores across different languages and datasets, without any language-specific tuning.
For Latin, a language with rich morphology, our methods even establish a new SOTA in Subtask 2 of SemEval'20 Task 1.

These results indicate that grammatical profiling
cannot compete with state-of-the-art methods based on large pre-trained language models, since they have the potential to encode both semantics and grammar. Yet reaching the highest possible scores on the task was not our goal. Instead, the aim of our study was to demonstrate that more attention should be paid to the relation between morphosyntax and semantic change. \new{Whether morphosyntactic and semantic features are complementary and can be successfully combined is a interesting question to be addressed in future work.}

We performed an extensive quantitative and qualitative analysis of our semantic change detection methods, showing that profiling yields interpretable results across several languages. Nevertheless, we still lack an understanding of some aspects of the interaction between semantics and morphosyntax. Finding the reasons behind the relatively
poor performance on some datasets, e.g.\ German, is an important direction for future studies.

Another interesting question is how to incorporate full dependency trees into grammatical profiles, rather than only \new{dependency relations to the syntactic head of a word}.
\newer{This is particularly important for analytical languages, where grammatical markers are presented in more than one word, such as with English verb mood and aspect. Moreover, dependency structure can be crucial for languages from families other than the Indo-European, e.g. to take into account detached counters in Japanese or plural markers in Yoruba.}

In light of our experimental results, we argue that grammatical profiling should become one of the standard baselines for semantic change detection.

\section*{Acknowledgements}
We thank the anonymous CoNLL-2021 reviewers for their helpful comments. This work has been partly supported by the European Union’s Horizon 2020 research and innovation programme under grants 770299 (NewsEye), 825153 (EMBEDDIA), and 819455 (DREAM).

\bibliography{anthology,our}
\bibliographystyle{acl_natbib}
\newpage
\appendix
\onecolumn

\section*{Appendix}
\label{sec:appendix}

\section{Model predictions}
\label{sec:app-predictions}
Table~\ref{tab:top10} shows the top 10 ranked words for each of the target languages according to the semantic change score of our best model. Because three time bins are available for Russian, we show the change score estimated for the interval between first and second (1-2), second and third (2-3), as well as first and third (1-3) periods.
\begin{table*}[h!]
\centering
\small
\begin{tabular}{l|l|l|l|l|l|l|l}
\toprule
\textbf{English} & \textbf{German} & \textbf{Latin} & \textbf{Swedish} & \textbf{Italian} & \textbf{Russian 1-2} & \textbf{Russian 2-3} & \textbf{Russian 1-3} \\ \midrule
gas              & Lyzeum          & imperator      & bröllop          & lucciola         & blagodarnost        & polosa               & ambitsia              \\
chairman         & vorweisen       & beatus         & studie           & palmare          & vek                  & vek                  & nalozhenie            \\
rag              & Schmiere        & regnum         & motiv            & tac              & sobrat               & zhest'                & ponedelnik          \\
stab             & zersetzen       & jus            & krita            & unico            & vyzov                & favorit              & vyzov                \\
ball             & verbauen        & adsumo         & konduktör        & pacchetto        & brat                 & sobrat               & blin                 \\
lass             & Eintagsfliege   & potestas       & annandag         & cappuccio        & jubiley               & ambitsia              & polosa               \\
prop             & beimischen      & licet          & aktiv            & egemonizzare     & ambitsia              & nalozhenie            & khren                 \\
tip              & Engpaß          & sensus         & granskare        & brama            & khren                 & jubiley               & uglevodorod          \\
record           & artikulieren    & nobilitas      & bolagsstämma     & campanello       & uglevodorod          & lishenie              & lishenie              \\
plane            & voranstellen    & sacramentum    & färg             & piovra           & ponedelnik          & blin                 & chastitsa   \\ \bottomrule
\end{tabular}
\caption{The top 10 rankings obtained with our best method for all the target languages. The topmost word is the one with the highest assigned change score. Russian words are transliterated from Cyrillic to Latin script.}
\label{tab:top10}
\end{table*}

\end{document}